# Stochastic Channel Decorrelation Network and Its Application to Visual Tracking

**Jie Guo,**[1] **Tingfa Xu,**[1] **Shenwang Jiang,**[1] **Ziyi Shen**[1]

[1]School of Optics and Photonics, Beijing Institute of Technology, China
20130261@bit.edu.cn, ciom_xtf1@bit.edu.cn, 3120160304@bit.edu.cn, joanshen0508@gmail.com

**Abstract**

Deep convolutional neural networks (CNNs) have dominated many computer vision domains because of their great power to extract good features automatically. However, many deep CNNs-based computer vison tasks suffer from lack of training data while there are millions of parameters in the deep models. Obviously, these two biphase violation facts will result in parameter redundancy of many poorly designed deep CNNs. Therefore, we look deep into the existing CNNs and find that the redundancy of network parameters comes from the correlation between features in different channels within a convolutional layer. To solve this problem, we propose the stochastic channel decorrelation (SCD) block which, in every iteration, randomly selects multiple pairs of channels within a convolutional layer and calculates their normalized cross correlation (NCC). Then a squared max-margin loss is proposed as the objective of SCD to suppress correlation and keep diversity between channels explicitly. The proposed SCD is very flexible and can be applied to any current existing CNN models simply. Based on the SCD and the Fully-Convolutional Siamese Networks, we proposed a visual tracking algorithm to verify the effectiveness of SCD.

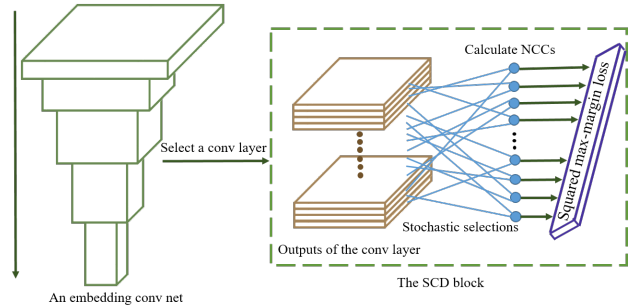

Figure 1: Left is an embedding conv net and right is the proposed stochastic channel decorrelation block.

## Introduction

Feature representation plays a key role in most computer vision tasks. Many works focus on designing or using robust features to improve the performance of their methods. Take visual tracking as an example. The research of (Wang et al. 2015) decomposes the visual tracking system into five parts, namely, motion model, feature extractor, observation model, model updater, and ensemble post-processor. They find that feature extractor is the most important component of a tracker. In their validation, the best feature combination (HOG + raw color) outperforms the basic model (raw grayscale) by more than 20%. The work in (Kalal, Mikolajczyk, and Matas 2012) uses local binary patterns as feature representation of the model. Histograms are popular features in vision community (Zhong, Lu, and Yang 2012; He et al. 2013). Haar-like features are another set of popular features for many vision tasks (Grabner, Grabner, and Bischof 2006; Babenko, Yang, and Belongie 2009; Hare, Saffari, and Torr 2015). Some models (Henriques et al. 2015; Danelljan et al. 2014a) take advantages of HOG descriptors. Researches in (Danelljan 2014b; Possegger, Mauthner, and Bischof 2015; Bertinetto et al. 2016a) use color features to account for target deformation. By leveraging rich gradient information across multiple color channels, the MC-HOG feature descriptor (Zhu et al. 2016) is proposed for handling drift problem. However, all the features mentioned above are hand-crafted features which are only designed for certain scenarios. For example, HOG features are stable when the target has abundant texture, while if the target has smooth appearance (e.g., infrared imagery), the feature may fail to accomplish the task. Thus, the hand-crafted features cannot qualify for the tracking of all generic objects. Besides, the hand-crafted features are incapable t o capture the sematic information which is significant for high-level vision tasks.

Recent years have witnessed the great representation power of deep Convolutional Neural Networks (CNNs) (Krizhevsky, Sutskever, and Hinton 2012; Chatfield et al. 2014, and He et al. 2016) trained on large scale image classification datasets (Russakovsky 2015). CNNs can learn to automatically abstract different levels (from low-level local features to high-level semantic features) of features. These features are generic enough to represent any objects if enough training data are given. However, many CNNs contain millions of parameters. Besides, some vision tasks, such as visual tracking, have very limited training data. These facts will result in parameter redundancy of many poorly designed deep CNNs, making the model overfit to the limited training data and lose generalization ability. Typical methods to solve these problems include transfer learning, increasing training data, shrinking the model, and adding regularization terms.

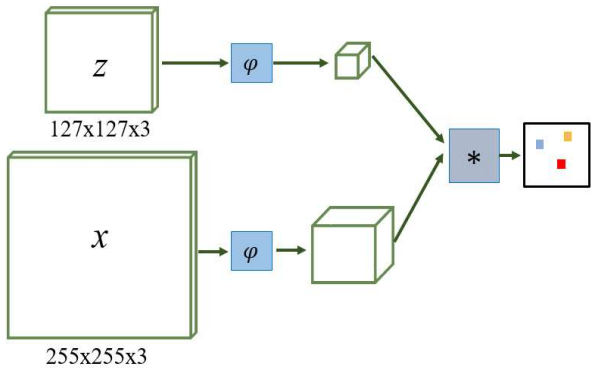

Figure 2: The architecture of the fully-convolutional Siamese tracker (SiamFC). The inputs are an exemplar image and a search image. They are passed through the same embedding conv net $\varphi$. The final score map is obtained by using a cross-correlation between the two output conv feature maps. The score map is scalar-valued with bigger value indicating more possibility that the object exists in the corresponding sub-window.

However, we look deep into the existing CNNs and find that the redundancy of network parameters comes from the correlation between features in different channels within a convolutional layer. To reduce the redundancy and maintain the diversity between features, we propose a stochastic channel decorrelation (SCD) method that can be applied to any conv layer of any current existing CNNs as shown in Fig. 1. The SCD also can be viewed as a regularization term of conv layers. We find that SCD is very effective even if the training data are rare. Applying SCD to the Baseline-conv5 tracker (Valmadre et al. 2017), we propose dozens of tracking algorithms to verify the effectiveness of SCD.

The contributions of this paper are listed as follows:
1. A novel stochastic channel decorrelation block is proposed to remove redundancy and keep diversity between channels within a conv layer.
2. Tracking algorithms are proposed based on SCD to verify the effectiveness of SCD. Experiments show the superiority of these tracking algorithms against other state-of-the-art trackers.
3. We do lots of experiments for analyzing different parts of SCD model, giving guides of choosing which layers to be decorrelated by SCD and which value of the margin of the proposed squared max-margin loss should be chosen.
4. The fine-tuning of models with SCD units is quite efficient and effective. We only fine-tune the model with limited video data (less than four hundred videos) and limited iterations (5000 iterations) to find that the method improves the performance of the baseline by a large margin without increasing any computational cost.

## Related Work

Many researches have been exploited to regularize the parameters of deep models and maintain diversity when training the models. Dropout (Srivastava et al. 2014) is a regularization method to prevent overfitting and co-adaptations when limited training data were given. It sets a subset of outputs within each fully connected layer to zeros randomly in every iteration. This equals to train a sub-network with much less parameters, thus alleviating overfitting. At test time, the test data are passed through the whole network to achieve the average predictions of all these sub-networks. Unlike Dropout, DropConnect (Wan et al. 2013) randomly turns off a subset of connections rather than just the output units, making it the generalization of Dropout. On the other hand, to improve the accuracy in human joint location estimation, SpatialDropout (Tompson et al. 2015) regularizes the convolutional layers by dropping a subset of channels randomly in convolutional layers. In visual tracking, BranchOut (Han, Sim, and Adam 2016) tries to develop an ensemble tracking algorithm by stochastic training multiple branches. Each base tracker is composed of a base conv net and a branch. The method randomly selects a subset of branches for model updating and thus keeps diversity between different branches. Similar to BranchOut, STCT (Wang et al. 2016) treats the tracking model as an ensemble of different individual base learners. Each base learner is trained using different loss criterions to reduce correlation and avoid over-training. All these methods tend to maintain model diversity and reduce correlation and co-adaptations of network parameters. They all try to achieve these goals by decomposing the model into multiple sub-models and individually training these sub-models with different diverse training data. Therefore, all sub-models are learned to qualify for the whole task individually without relying on others, thus reducing co-adaptations while keeping diversity. Our model, on the contrary, does not need to separately train a subset of parameters in every iteration. It can keep the model as a whole while still achieving diversity by explicitly telling the network to reduce the correlation between the learned features through the proposed SCD block. The usage of SCD is quite flexible. It can be applied to any conv layer of any existing CNN model.

Table 1: Architecture of convolutional embedding function $\varphi$ of the Baseline.

| Layer | Kernel size | Stride | Activation size for exemplar | Activation size for search | Chans. |
|---|---|---|---|---|---|
|  |  |  | 127×127 | 255×255 | 3 |
| conv1 | 11×11 | 2 | 59×59 | 123×123 | 96 |
| pool1 | 3×3 | 2 | 29×29 | 61×61 | 96 |
| conv2 | 5×5 | 1 | 25×25 | 57×57 | 256 |
| pool2 | 3×3 | 1 | 23×23 | 55×55 | 256 |
| conv3 | 3×3 | 1 | 21×21 | 53×53 | 384 |
| conv4 | 3×3 | 1 | 19×19 | 51×51 | 384 |
| conv5 | 3×3 | 1 | 17×17 | 49×49 | 32 |

Since we apply SCD to the Baseline-conv5 model (Valmadre et al. 2017) and evaluate SCD in visual tracking task, we illustrate the Baseline-conv5 model in detail. In the following text, we will use Baseline instead of Baseline-conv5 for short. The Baseline is an updated version of SiamFC (Bertinetto et al. 2016b). It differs from SiamFC in two ways. Firstly, it uses less strides (total 4 instead of 8 strides) in the network to get lager features maps. This is good for our model because lager size features are easier to decorrelate than small size features. Secondly, final layer has only 32 instead of 256 output channels in order to preserve the high speed processing. The two changes do not influence the tracking performance of SiamFC. The Baseline and SiamFC is based on Siamese networks (Chopra, Hadsell, and LeCun, 2015) which have advantages of learning matching functions by sharing weights between two branches. Their architectures is fully-convolutional with respect to the search image. Fig. 2 shows the network structure of them. It takes a template and a larger search image as inputs. The final score map is obtained by using a cross-correlation between the two output conv feature maps. The score map is scalar-valued with bigger value indicating more possibility that the object exists in the corresponding sub-window. In every frame, it just need to pass the image pair through the network. Then the target will be located. The method is very efficient that it runs at speeds that far exceed the real-time requirement. However, despite the high speed, the performance is not that convincing compared with many other state-of-the-art trackers.

Obviously, diverse features are good to deal with numerous tracking challenges (illumination change, deformation, motion blur, low resolution) in visual tracking. Besides, with limited training data and vase redundant network parameters, the model may overfit to the data it has seen. These shortages may be the key reasons causing performance degradation. To deal with these problems, we apply the SCD block to the conv layers of the Baseline and help the conv features maintain diversity and avoid correlation explicitly. The network structure of the Baseline model is illustrated in Table 1. Other details of the Baseline and SiamFC trackers please refer to (Valmadre et al. 2017; Bertinetto et al. 2016b).

## Stochastic Channel Decorelation

The SCD block is a non-parameter unit which can be applied to any conv layer. It achieves the goal of maintaining conv feature diversity by explicitly penalizing the correlation between conv features using a squared max-margin loss function.

**Preliminaries and motivations:** Denote $I \in \mathbb{R}^{H' \times W' \times C'}$ as the input features of a conv layer, where $H'$, $W'$ and $C'$ represent the height, width and channel sizes of the input features, respectively. Denote $O \in \mathbb{R}^{H \times W \times C}$ as the output features of the conv layer. Let $K = [k_1, k_2, ..., k_C]$ represent the set of filter kernels of this conv layer, where $k_c$ refers to the $c^{th}$ filter. Then we can obtain the output features $O = [o_1, o_2, ..., o_C]$ by

$$o_c = k_c * I = \sum_{c'=1}^{C'} k_c^{c'} * i^{c'}, \quad (1)$$

where $*$ denotes convolution, $k_c = [k_c^1, k_c^2, ..., k_c^{C'}]$ and $I = [i^1, i^2, ..., i^{C'}]$. Note that, for simplicity, we omit the bias terms.

A conv layer that gives good representation of general objects should meet two requirements. Firstly, different channels in the input features I of the conv layer should be diverse enough. Since if the input features (i.e., $i^1, i^2, ..., i^{C'}$) within I are correlated with each other, then there is no need to assign an independent kernel $k_c^{c'}$ to each input features. Because if do so, the parameters (i.e., $k_c^1, k_c^2, ..., k_c^{C'}$) within $k_c$ will be redundant, thus causing overfitting. Secondly, different channels in the output features O should be diverse enough. Since we want the conv layer to learn more diverse features. Besides, the inputs of all filters are the same (i.e., I ) and if the output features (i.e., $o_1, o_2, ..., o_C$) of all the filters are correlated with each other, it indicates that the parameters (i.e., $k_1, k_2, ..., k_C$) within K are highly correlated and redundant. Thus the need to suppress the correlation between feature channels is urgent. But, how to accomplish this goal? Naturally and intuitively, we compute the correlation degree of two features which are from any two different channels of a conv layer and minimize it. For this reason, we propose the SCD block. Since the output features O of a conv layer is the input features I of the next conv layer, we only need to apply the SCD block to the output features of each conv layer.

**Correlation measurement:** We use the normalized cross correlation (NCC) as the measurement of the correlation degree of any channel pair,

$$p_{m,n} = \frac{\sum o_m \cdot o_n}{\sqrt{\sum o_m \cdot o_m \times \sum o_n \cdot o_n}}, \quad (2)$$

where $\cdot$ denotes element-wise multiplication, $p_{m,n}$ represents the correlation degree between feature channel $o_m$ and $o_n$, and $m, n \in [1, 2, ..., C], \text{s.t.} m \neq n$. The NCC has some good properties. Firstly, it has been widely used (Kalal, Mikolajczyk, and Matas 2012; Tao, Gavves, and Smeulders 2016) as a similarity function of two features and proved to be effective. Secondly, the value of NCC ranges from -1 to 1 which can be directly used to indicate the correlation degree of any two features.

**Squared max-margin loss:** Now we obtain the correlation degree of two feature channels. The original goal is to suppress the correlation and maintain diversity. Hence, we propose a squared max-margin loss to achieve this goal by,

$$\mathcal{L}_{m,n} = \max\left(0, \ \text{abs}(p_{m,n}) - \epsilon\right)^2, \quad (3)$$

where $\text{abs}(p_{m,n})$ is the absolute value of $p_{m,n}$ and $\epsilon$ is the margin value. The value of $\epsilon$ has strong influence on the performance of the model. If $\epsilon$ is too big, for example, larger than 0.5, the SCD may not work, since in very rare cases $\text{abs}(p_{m,n})$ will be larger than 0.5. On the other hand, if $\epsilon$ is too small, for example 0, the model will force all channels to be totally uncorrelated with each other. This is also very irrational, because sometimes different features share parts of good features will help improve the performance. Besides, if the dimension of feature $o_c$ is smaller than the total number of channels in O, it is impossible to let all feature channels be totally uncorrelated with each other. The experimental comparisons of models using different values of $\epsilon$ are discussed in the experiment section in detail.

**Stochastic channel decorrelation:** Note that Eq. (3) is only the loss of one pair of features. Nevertheless, we intend to suppress the correlation of any pair of features. Hence, intuitively, we need to scan through the entire pairs of features in the output features of one layer. However, this ergodic operation will cause big computational burden, especially when the number of channel in the output features is large. Motivated by the stochastic gradient descent (SGD) algorithm, we randomly select a subset of feature pairs in every iteration. Over time, all pairs of features will be taken into consideration. Let set $U = \{a, b | a, b \in [1, 2, ..., C], a \neq b\}$. We randomly select $M$ elements in U, forming the selected set $S = \{s_1, s_2, ..., s_M\}$. Then the loss function of the corresponding conv layer in this iteration is,

$$\mathcal{L}^l = \frac{1}{M} \sum_{m=1}^{M} \mathcal{L}_{s_m}, \quad (4)$$

where $\mathcal{L}^l$ denotes the loss of the $l^{th}$ conv layer. In our experiment the size of the subset feature pairs $M$ is restricted to 1000 if the total number of pairs is larger than 1000. This number is chosen to meet a balance between computational burden and integrality.

**The selection of conv layers to be decorrelated by SCD:** Features from different conv layers catch different kinds of information and instance property. The features in the early layers retain low-level visual information, which is beneficial for visual task problems like motion blur and low resolution. It would help alleviate these problems if we apply SCD to the early conv layers to maintain the diversity of low-level information. On the other hand, the features in the latter layers capture more semantic information, which is good for problems like background clutter, deformation, and occlusion. We can solve these problems by applying SCD to latter layers. Denote $L = \{l_1, l_2, ..., l_N\}$ as the layer number set, which labels the conv layers to be decorrelated by SCD. Here, $N$ is the total number of conv layers being decorrelated. Then the final loss function of the whole model is given by,

$$\mathcal{L} = \alpha \mathcal{L}_{task} + \beta \frac{1}{N} \sum_{n=1}^{N} \mathcal{L}^{l_n}, \quad (5)$$

where $\mathcal{L}_{task}$ is the loss function of the visual task that the model need to fulfil, the term $\frac{1}{N} \sum_{n=1}^{N} \mathcal{L}^{l_n}$ represents the total loss of all the SCD units. Parameters $\alpha$ and $\beta$ are to balance the losses between the visual task and the SCD units. In our experiment, we set $\alpha = 1$ and $\beta = 2$ empirically.

**Implementation details:** We utilize the implementation of Baseline-conv5 (Valmadre et al. 2017) which is a variant of SiamFC (Bertinetto et al. 2016b) as baseline of our work as illustrated in the related work section. Then we apply SCD to the Baseline model, producing dozens of models which use different values of $\epsilon$ and select different conv layers to be uncorrelated by SCD. All these models are fine-tuned by the same training settings. The size of mini-batch is 32. 100 samples are extracted from each training sequence in each epoch and each model is fine-tuned for totally 50 epochs. The initial learning rate is set to 0.001. After each epoch, the learning rate is multiplied by a fixed factor until reaching 0.00001 in the last epoch. The models are trained by SGD algorithm with weight decay 0.0005. Note that the SCD block is applied before the batch normalization (Ioffe and Szegedy 2015) layer.

## Experiment

We use the sequences in ALOV (Smeulders et al. 2014) to fine-tune all the models. Note that we exclude the 14 sequences that overlap with our test sets OTB2013 (Wu, Lim, and Yang 2013), OTB100 (Wu, Lim, and Yang 2015), and OTB50 (Wu, Lim, and Yang 2015). We name all variants of our trackers as the format of SCDT_L$a$_M$b$, where $a\epsilon\{1,2,3,4,5,12,13,35,45,135,12345\}$ denotes the layer/layers to be decorrelated by SCD and $b \in \{0.0, 0.1, 0.2, 0.3, 0.4\}$ represents the margin value $\epsilon$ in Eq. (3). To make the experiments more convincing, we also fine-tune the Baseline model using the training dataset without applying any SCD unit. We name the corresponding tracker SCDT_none. This is to certify that the performance gain of the proposed models does not benefit from the extra fine-tuning dataset. We run all the models in a single NVIDIA Titan X GPU and all the trackers run at speed of 54 fps (frames per second) which far exceeds the real-time requirement. The best variant of our models improves the performance of the Baseline model by a large margin without increasing any computational cost.

Table 2: Performance as overlap (IoU) and precision produced by the OTB toolkit for the OTB2013, OTB100, and OTB50 datasets. The best three results are shown in **red**, **blue**, and **green** bold fonts. Best viewed in color.

|  | OTB2013 | | OTB100 | | OTB50 | |
|---|---|---|---|---|---|---|
| Method | IoU | Prec. | IoU | Prec. | IoU | Prec. |
| SCDT_L35_M0.3 | **0.659** | **0.894** | **0.620** | **0.836** | **0.579** | **0.786** |
| SCDT_L3_M0.3 | **0.646** | **0.867** | **0.609** | **0.816** | **0.564** | **0.757** |
| SCDT_L5_M0.3 | 0.642 | 0.860 | **0.603** | **0.810** | **0.565** | 0.751 |
| SCDT_L13_M0.3 | 0.636 | 0.865 | 0.602 | 0.809 | 0.553 | 0.744 |
| SCDT_L45_M0.3 | **0.647** | **0.870** | 0.601 | 0.808 | 0.552 | 0.742 |
| SCDT_L2_M0.3 | 0.630 | 0.837 | 0.598 | 0.805 | 0.552 | 0.739 |
| SCDT_L12345_M0.3 | 0.635 | 0.855 | 0.598 | 0.802 | 0.561 | **0.764** |
| SCDT_L1_M0.3 | 0.621 | 0.849 | 0.597 | 0.800 | 0.539 | 0.737 |
| SCDT_L135_M0.3 | 0.631 | 0.845 | 0.594 | 0.800 | 0.544 | 0.728 |
| SCDT_L12_M0.3 | 0.616 | 0.839 | 0.592 | 0.799 | 0.534 | 0.736 |
| SCDT_L4_M0.3 | 0.626 | 0.841 | 0.590 | 0.789 | 0.532 | 0.710 |
| Baseline | 0.618 | 0.806 | 0.588 | 0.771 | 0.517 | 0.683 |
| SCDT_none | 0.606 | 0.806 | 0.571 | 0.766 | 0.520 | 0.702 |
| SiamFC3s | 0.607 | 0.809 | 0.582 | 0.769 | 0.516 | 0.692 |

Two evaluation metrics are employed in our experiment: bounding box overlap ratio and center location error in the one-pass evaluation (OPE) (Wu, Lim, and Yang 2013) protocol.

**Different conv layer/layers with SCD:** Table 2 shows the performance of trackers with different conv layers being decorrelated by SCD. From the table we draw four conclusions: Firstly, all the network models decorrelated by SCD have better performance than the base line model. Especially, the best variant SCDT_L35_M0.3 improves the performance of the Baseline model by a large margin in terms of all the three datasets. This verifies the effectiveness of SCD. Secondly, the best three variants among these trackers are SCDT_L35_M0.3 SCDT_L3_M0.5, and SCDT_L5_M0.3 which apply SCD to only one or two layer/layers, while the SCDT_L12345_M0.3 tracker which applies SCD to all its conv layers has mediocre performance. This illustrates that more conv layers being decorrelated may not improve the performance any more. Thirdly, latter layers being decorrelated is more effective than early layers being decorrelated. Fourthly, The SCDT_none tracker does not improved by only fine-tuning the Baseline model, which states that the performance gain of all the SCDTs does not come from the additional training data. It is the SCD block that matters.

**Different scenarios with SCD:** Table 3 illustrates the performance of different trackers in terms of five different tracking problems (i.e., motion blur, low resolution, background clutter, deformation, and occlusion) to further help understand what benefits can we get by applying SCD to different conv layers. For simplicity, Table 3 only displays the overlap (IoU) scores of OTB2013 dataset. More experiment results please refer to the *supplemental materials*. In OTB2013 dataset, there are 12 sequences with motion blur, 4 sequences with low resolution, 21 sequences with background clutter, 19 sequences with deformation, and 29 sequences with occlusion. We run the trackers on these sequences which have different attributes separately. The results are displayed in Table 3.

The upper rows in Table 3 are the trackers with the early conv layers being decorrelated by SCD while the lower rows in Table 3 are the trackers with the latter conv layers being decorrelated. Motion blur and low resolution are tracking problems which need higher spatial resolution and local detail information for better performance. From Table 3 we know that the models with early conv layer/layers being decorrelated by SCD generally perform better. Since applying SCD to the early conv layers helps the model better exploit these low-level visual information. On the contrary, background clutter, deformation and occlusion are tracking problems which will benefit from high-level semantic information rather than low-lever visual information. Interestingly, the trackers with latter layer/layers, which capture more semantic information, decorrelated by SCD have better performance than others in these tracking scenarios.

Table 3: The IoU scores of the trackers evaluating with OTB2013 dataset in terms of five attributes. The best three results are shown in **red**, **blue**, and **green** bold fonts. Best viewed in color.

| Method | Motion blur | Low resolution | Background clutter | Deformation | Occlusion |
|---|---|---|---|---|---|
| SCDT_L12_M0.3 | **0.608** | **0.649** | 0.566 | 0.576 | 0.600 |
| SCDT_L13_M0.3 | 0.574 | 0.541 | 0.580 | **0.614** | 0.604 |
| SCDT_L1_M0.3 | 0.515 | 0.491 | 0.572 | 0.594 | 0.604 |
| SCDT_L2_M0.3 | 0.582 | **0.648** | 0.589 | 0.571 | 0.591 |
| SCDT_L12345_M0.3 | **0.600** | 0.637 | 0.605 | 0.571 | 0.600 |
| SCDT_L3_M0.3 | 0.564 | **0.646** | **0.616** | 0.601 | **0.613** |
| SCDT_L135_M0.3 | 0.528 | 0.501 | 0.579 | 0.602 | 0.606 |
| SCDT_L4_M0.3 | 0.513 | 0.478 | 0.575 | 0.583 | 0.592 |
| SCDT_L5_M0.3 | 0.555 | 0.579 | **0.607** | 0.593 | 0.609 |
| SCDT_L35_M0.3 | **0.613** | 0.641 | **0.617** | **0.615** | **0.636** |
| SCDT_L45_M0.3 | 0.569 | 0.445 | 0.604 | **0.626** | **0.635** |

Table 4: The IoU scores of the trackers evaluating with OTB2013 dataset in terms of different margin values. The best three results are shown in **red**, **blue**, and **green** bold fonts. Best viewed in color.

| Method | M0.0 | M0.1 | M0.2 | M0.3 | M0.4 |
|---|---|---|---|---|---|
| SCDT_L12 | **0.621** | 0.607 | **0.611** | **0.616** | 0.594 |
| SCDT_L13 | 0.618 | 0.617 | **0.619** | **0.636** | **0.633** |
| SCDT_L1 | 0.610 | **0.634** | 0.631 | 0.621 | **0.630** |
| SCDT_L2 | 0.620 | **0.631** | 0.630 | 0.630 | 0.623 |
| SCDT_L12345 | **0.631** | 0.614 | 0.617 | **0.635** | **0.629** |
| SCDT_L3 | 0.642 | **0.648** | 0.632 | **0.646** | **0.647** |
| SCDT_L135 | 0.611 | 0.617 | 0.625 | **0.631** | **0.630** |
| SCDT_L4 | **0.630** | 0.620 | **0.627** | 0.626 | **0.637** |
| SCDT_L5 | 0.631 | 0.638 | **0.639** | **0.642** | **0.653** |
| SCDT_L35 | 0.629 | 0.639 | **0.659** | **0.659** | **0.641** |
| SCDT_L45 | 0.628 | 0.635 | **0.657** | **0.647** | **0.642** |

To sum up, applying SCD to different conv layers will help better exploit the corresponding level of feature information. For low-level visual task, like image deblurring, demosaicking, and denoising, we suggest applying SCD to the early conv layers. Meanwhile, for high-level visual task, like object recognition, gait recognition, and semantic segmentation, we suggest applying SCD to the latter conv layers.

**The selection of margin value:** Table 4 shows the performance of different trackers in terms of different margin values. The first column specifies trackers with which conv layer/layers being decorrelated by SCD. The first row specifies the margin values used in each model. The trackers SCDT_L1, SCDT_L2 and SCDT_L12 perform the best when the margin values are 0.1, 0.1 and 0.0, respectively. All the three trackers apply SCD to the early layer/layers of the baseline model, indicating that early layer prefer smaller margin value. Meanwhile, SCDT_L5, SCDT_L4 and SCDT_L35 perform best when the margin values are 0.4, 0.4 and 0.3/0.2, respectively. These trackers apply SCD to their latter conv layer/layers, indicating that latter layer prefer lager margin value. Generally, the trackers with margin value 0.3 perform better than the trackers with other margin values. As illustrated in the *squared max-margin loss section*, if the margin value $\epsilon$ is too big, for example, larger than 0.5, the SCD may not work, since in very rare cases $\text{abs}(p_{m,n})$ will be larger than 0.5. On the other hand, if $\epsilon$ is too small, for example 0, the model will force all channels to be totally uncorrelated with each other. This is also very irrational, because sometimes different features share part of good feature will help improve the performance.

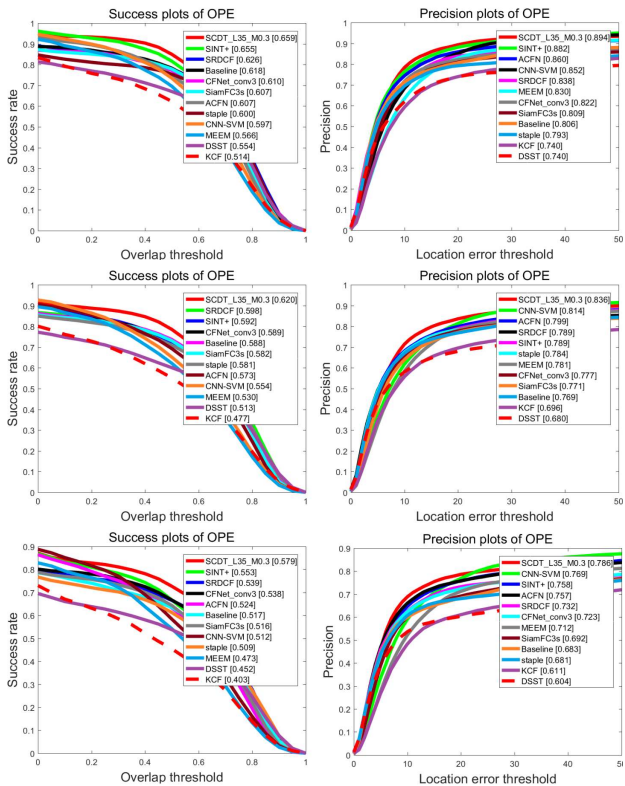

Figure 3: The first, second, and third rows represent the tracking results on OTB2013, OTB100, and OTB50, respectively. Best viewed in color.

We know that different conv layers may have different optimal margin values. However, it is impossible to conduct all the experiments for comparison since the combinations are too many. Developing a method to decide the value of margin in each SCD loss function automatically is necessary. However, this is outside the scope of this paper. In this paper, we set the margin values within each model to be the same for simplicity.

**Compare with the-state-of-the-art:** The best overall performance tracker among all the variants of our trackers is SCDT_L35_M0.3. Hence it is used to compare with other state-of-the-art trackers, including Baseline (Valmadre et al. 2017), CFNet_conv3 (Valmadre et al. 2017), SINT+ (Tao, Gavves, and Smeulders 2016), SiamFC3s (Bertinetto et al. 2016b), SRDCF (Danelljan et al. 2016), ACFN (Choi et al. 2017), staple (Bertinetto et al. 2016a), CNN-SVM (Hong et al. 2015), MEEM (Zhang, Ma, and Sclaroff 2014), DSST (Danelljan et al. 2014a), and KCF (Henriques et al. 2015). Fig. 3 compares the tracking results of these trackers.

The SCDT_L35_M0.3, Baseline, CFNet_conv3, SINT+, and SiamFC3s trackers are all based on the Siamese deep convolutional neural network architecture and SCDT_L35_M0.3 performs the best among these trackers, which indicates the proposed SCD block can significantly improve the Siamese convolutional network models. The SRDCF, ACFN, staple, DSST, and KCF trackers represents the methods based on correlation filters. The proposed method outperforms all other compared trackers in all the three datasets.

The proposed tracker does not increase any computational cost of the Baseline tracker and it runs at 54 fps (frames per second) which far exceeds the real-time requirement.

## Conclusion

Many deep CNNs based computer vison tasks suffer from lack of training data while the deep models may have millions of parameters. Obviously, these two biphase violation facts reflect the parameter redundancy of many poorly designed deep CNN models. To reduce the redundancy while still keep capability of the deep model, a stochastic channel decorrelation (SCD) block is proposed to suppress correlation and maintain diversity between features in different channels within a conv layer. The SCD block is quite simple and flexible but effective. It can be applied to any conv layer of any existing CNN model by rarely changing the architecture of the original model. It is also a non-parameter unit, which does not increase any computational cost of the original model.

To verify the effectiveness of SCD, we apply SCD block to the SiamFC model for visual tracking. Two key factors, i.e., which layer/layers to be decorrelated by SCD and what value of the margin in the loss function should be chosen, are discussed in detail with plentiful experiments. Finally, our method achieves favorably performance against other state-of-the-art trackers. Especially, our method outperforms the Baseline model by a large margin without reducing the running speed of the tracker.